\PassOptionsToPackage{table}{xcolor}  

\documentclass[10pt,twocolumn,letterpaper]{article}

\usepackage{iccv}              

\newcommand{\calC}{{\mathcal{C}}}

\newcommand{\calL}{{\mathcal{L}}}

\newcommand{\be}{\begin{eqnarray}}
\newcommand{\ee}{\end{eqnarray}}
\newcommand{\bee}{\begin{eqnarray*}}
\newcommand{\eee}{\end{eqnarray*}}

\newcommand{\matrixb}{\left[ \begin{array}}
\newcommand{\matrixe}{\end{array} \right]}

\newcommand{\Tref}[1]{Table~\ref{#1}}

\newcommand{\Fref}[1]{Fig.~\ref{#1}}

\newcommand{\Sref}[1]{Sec.~\ref{#1}}

\def\eg{\emph{e.g.}}

\def\etal{\emph{et al.}}
\def\ie{\emph{i.e.}}
\usepackage{dsfont}
\usepackage{makecell}
\definecolor{verylightgray}{gray}{0.92}
\newcommand{\mytabular}[2]{\centering\scalebox{#1}{#2}}
\usepackage[pagebackref,breaklinks,colorlinks,allcolors=iccvblue]{hyperref}
\definecolor{iccvblue}{rgb}{0.21,0.49,0.74}
\usepackage[pagebackref,breaklinks,colorlinks,allcolors=iccvblue]{hyperref}

\title{MambaVideo for Discrete Video Tokenization with Channel-Split Quantization}

\author{
Dawit Mureja Argaw$^{1,2}$\quad
Xian Liu$^{1}$\quad
Joon Son Chung$^2$ \quad
Ming-Yu Liu$^1$ \quad
Fitsum Reda$^1$
 \\ \\
$^1$ NVIDIA \quad $^2$  KAIST\\
}

\begin{document}
\maketitle
\begin{abstract}
Discrete video tokenization is essential for efficient autoregressive generative modeling due to the high dimensionality of video data. This work introduces a state-of-the-art discrete video tokenizer with two key contributions. First, we propose a novel Mamba-based encoder-decoder architecture that overcomes the limitations of previous sequence-based tokenizers. Second, we introduce a new quantization scheme, channel-split quantization, which significantly enhances the representational power of quantized latents while preserving the token count. Our model sets a new state-of-the-art, outperforming both causal 3D convolution-based and Transformer-based approaches across multiple datasets. Experimental results further demonstrate its robustness as a tokenizer for autoregressive video generation.
\end{abstract}
\vspace{-4mm}
\section{Introduction}
\label{sec:intro}
Discrete video tokenization (DVT) aims to map a video into a sequence of discrete representations for autoregressive generative modeling, addressing the curse of dimensionality inherent in video-related tasks. The most commonly used approaches for DVT follow three main steps: first, an \emph{encoder} network compresses the input video into latent features; second, a \emph{quantization} layer maps the continuous encoded features to discrete tokens (codes) using a codebook; and third, a \emph{decoder} network reconstructs the input from these discrete tokens. The effectiveness of DVT largely depends on two key components: (1) the encoder-decoder architecture, which governs the overall compression and reconstruction process, and (2) the quantization mechanism, which determines the quality and efficiency of discrete representations.

Existing discrete video tokenizers broadly fall into two categories based on their encoder-decoder architectures: 3D convolution-based~\cite{yan2021videogpt,yu2023language} and Transformer-based~\cite{villegas2022phenaki,wang2024omnitokenizer}. While state-of-the-art discrete image tokenizers~\cite{yu2021vector,huang2023towards} use vision Transformers (ViTs)~\cite{dosovitskiy2020image}, leading video tokenizers~\cite{yu2023language,gupta2023photorealistic} favor causal 3D convolutions for their superior computational efficiency with video data. The reliance on positional embeddings in Transformer-based tokenizers also makes it difficult to tokenize unseen spatial and temporal resolutions, as noted in Yu~\etal~\cite{yu2023language}. Furthermore, unlike 3D convolution-based tokenizers that employ hierarchical encoding and decoding, Transformer-based tokenizers~\cite{villegas2022phenaki,wang2024omnitokenizer} rely on single \emph{patchify} and \emph{topixel} layers to directly downsample the input video to the target latent dimension and reconstruct it, respectively, which limits spatio-temporal attention to a fixed latent size.

To overcome the limitations of previous sequence-based tokenizers, we propose a novel encoder-decoder architecture for DVT. Our model employs a hierarchical framework, where the encoder network downscales the input video in a top-down manner through a series of encoder blocks, each consisting of a cascade of \emph{patchify} and spatial-temporal attention modules. Similarly, the decoder upscales the quantized latent in a bottom-up fashion using a series of decoder blocks, each containing a cascade of attention and \emph{topixel} modules (see \Fref{fig:arch_overview}c). Unlike previous works~\cite{villegas2022phenaki, wang2024omnitokenizer} that use a linear embedding layer in the \emph{patchify} and \emph{topixel} modules, our model employs a 3D convolution-based embedding layer to better capture dependencies between spatio-temporal patches. 

To further enhance encoding and decoding, we introduce residual connections within the encoder and decoder blocks via token pooling and interpolation, respectively. 
Additionally, we utilize Mamba~\cite{gu2023mamba, dao2024transformers} layers instead of Transformers~\cite{vaswani2017attention} for the spatial and temporal attention modules. This choice is motivated by the fact that Mamba is a powerful model for reasoning over long-sequence inputs and \emph{does not require explicit positional encoding}, as it operates in a recurrent manner. As a result, it effectively mitigates the positional encoding bias that limits the generalization capabilities of Transformer-based tokenizers~\cite{villegas2022phenaki}. Furthermore, Mamba's linear-scale attention significantly enhances computational efficiency compared to Transformers, enabling high-resolution training and inference.

Beyond the encoder-decoder framework, quantization plays a critical role in DVT. Most discrete tokenization methods~\cite{van2017neural,esser2021taming,villegas2022phenaki,wang2024omnitokenizer} use vector quantization (VQ)~\cite{gray1984vector}, which relies on a learnable codebook optimized for compressed, semantic data representation. However, VQ presents several challenges: training the codebook is unstable and requires extra losses and hyperparameters~\cite{van2017neural,esser2021taming}; larger codebooks are frequently underutilized, hurting generative performance~\cite{mentzer2023finite,yu2023language}; and it is computationally inefficient due to the need to search through all codebook entries to find the closest match to the encoder output.

To address these challenges, recent works~\cite{yu2023language,mentzer2023finite,zhao2024image} have explored quantization schemes based on non-learnable codebooks. For example, Yu~\etal~\cite{yu2023language} introduced look-up free quantization (LFQ), which transforms latent values in the channel dimension into a binary sequence of -1's and 1's. Similarly, Mentzer~\etal~\cite{mentzer2023finite} proposed finite-scalar quantization (FSQ), which quantizes latent values to a fixed set, forming an implicit codebook generated by the product of these sets. While these approaches alleviate VQ’s limitations, their quantized latents suffer from limited \emph{representational power}. For instance, LFQ restricts latent values to binary representations, whereas VQ allows real-valued representations. FSQ, on the other hand, requires a much smaller latent dimension to maintain non-overlapping mappings, which limits its flexibility compared to VQ. As a result, both LFQ and FSQ-based tokenizers heavily rely on the decoder network for reconstruction, which constrains their overall generalization ability.

To mitigate this challenge, we introduce a new quantization scheme, termed \emph{channel-split quantization}, which enhances the representational power of the quantized latent while preserving the number of tokens and can be easily integrated into both LFQ and FSQ. The key idea is to leverage the trade-off between spatio-temporal compression and the quantization steps. Let $c$ denote the required channel dimension for the base quantizer,~\eg~FSQ. First, the encoder produces a latent representation with a channel dimension of $c \cdot K$, where $K > 1$. The latent is then \emph{split} into $K$ groups along the channel dimension, and each split is quantized independently using FSQ. The resulting latents are concatenated channel-wise and passed to the decoder, a processes we refer to as \emph{channel-split FSQ (CS-FSQ)}. To maintain the same number of quantized tokens as naive FSQ, we offset the increased channel dimension by increasing the encoder's spatio-temporal compression rate by a factor of $K$.

Unlike LFQ/FSQ, where each pixel in the encoded latent is represented by a \emph{single} token, channel-split LFQ/FSQ represents each pixel as a \emph{permutation-sensitive} sequence of tokens, thereby enhancing its representational capability. Given a codebook size of $2^N$, we prove theoretically that channel-split quantization effectively increases representation capacity (achieving an effective codebook size $\gg 2^N$) while maintaining the number of tokens (refer to~\Sref{sec:wh_cs_works}). 

By coupling the proposed Mamba-based architecture with our channel-split quantization scheme, we introduce a new discrete video tokenizer. We evaluate our method against state-of-the-art approaches~\cite{yan2021videogpt,villegas2022phenaki,wang2024omnitokenizer,yu2023language} on different video benchmarks~\cite{pont20172017,Niklaus_CVPR_2020}. Our experimental results demonstrate that our model establishes a new state-of-the-art in video tokenization. Furthermore, we integrate our pretrained tokenizer into an open-source autoregressive framework~\cite{yan2021videogpt} and train it for unconditional video generation on the SkyTimelapse~\cite{dtvnet} and UCF-101~\cite{soomro2012ucf101} datasets. The results strongly validate our model as a robust tokenizer for training autoregressive video generation models.
\vspace{-2mm}
\section{Proposed Encoder--Decoder Architecture}
\label{sec:method_enc_dec}
\begin{figure*}[!t]
    \centering
    \includegraphics[width=1\linewidth, trim={0.65cm, 3.45cm, 6.9cm, 0.95cm}, clip]{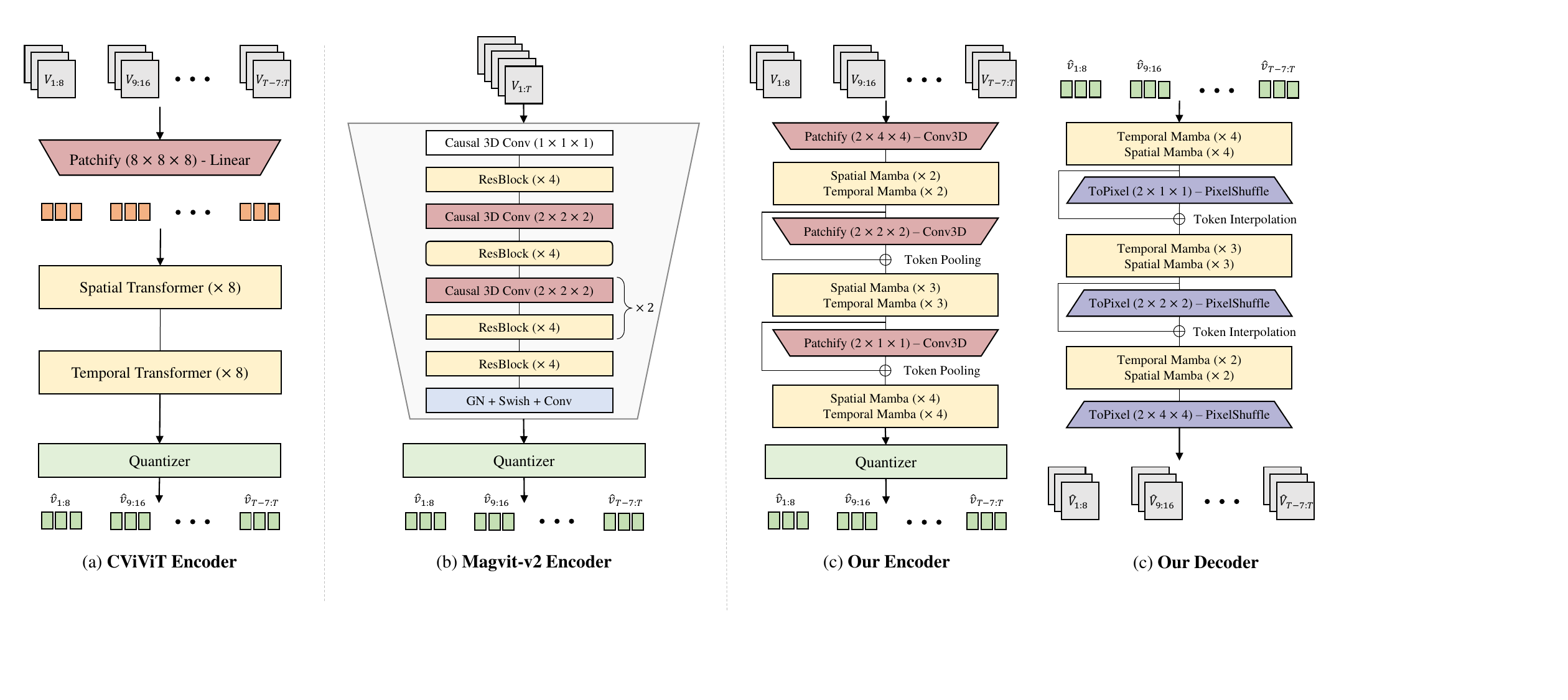}
    \caption{\textbf{Architecture Overview}: (a) The encoder network for CViViT~\citep{villegas2022phenaki}, a state-of-the-art Transformer-based tokenizer. (b) The encoder network for Magvit-v2~\citep{yu2023language}, a state-of-the-art causal 3D convolution-based tokenizer (c) The encoder and decoder architecture of the proposed Mamba-based tokenizer. Each model is designed with an $8 \times 8 \times 8$ spatio-temporal compression rate.}
    \label{fig:arch_overview}
    \vspace{-5mm}
\end{figure*}

\vspace{-1mm}
Our work aims to develop a robust sequence-based discrete video tokenizer for autoregressive generative modeling. The key idea is to enhance sequence-based tokenizers with hierarchical encoding and decoding, leveraging 3D convolution-based \emph{patchify} and \emph{topixel} layers, residual connections through token pooling and interpolation, and efficient spatio-temporal attention using Mamba layers. An overview of the proposed network is illustrated in~\Fref{fig:arch_overview}c.
\vspace{-1mm}
\subsection{Encoder}
Given a video $V$ of size $T \times H \times W \times 3$ and a target spatio-temporal compression rate of $\times thw$, previous Transformer-based works~\cite{villegas2022phenaki,zhao2024image,wang2024omnitokenizer} employ a single \emph{patchify} layer (\ie~a kernel of size $t \times h \times w $) to downsample the video into a feature $v$ of size $ T/t \times H/h \times W/w \times c$. The feature $v$ is then processed by a cascade of spatial and temporal attention modules to obtain the encoded latent representation (see \Fref{fig:arch_overview}a). However, this design constrains spatio-temporal reasoning to a fixed latent size, unlike the hierarchical encoding used in 3D convolution-based tokenizers~\cite{yu2023language} (see \Fref{fig:arch_overview}b). This limitation leads to inferior performance, especially at higher compression rates~\cite{yu2023language}. To address this issue, we implement hierarchical spatio-temporal downsampling in a top-down manner using a series of encoder blocks. Each block consists of a cascade of \emph{patchify}, \emph{spatial attention}, and causal \emph{temporal attention} modules, as depicted in~\Fref{fig:arch_overview}c.
\vspace{-4mm}
\paragraph{Patchify} The \emph{patchify} module reduces the dimensions of a given video both spatially and temporally. It contains a \emph{reshape} layer that rearranges an input visual data into a sequence of 
spatio-temporal patches (tokens) and an \emph{embedding} layer that extracts a feature representation from each patch. Let $L$ denote the total number of levels (blocks) in the encoder. The \emph{patchify} module at each level $l$, where $l \in [1,L]$, downsamples the input feature with a spatio-temporal kernel of size $t_l \times h_l \times w_l$. This process is repeated at each encoder block and the final output of the encoder has a dimension $T/t \times H/h \times W/w \times c$, where $t = \prod_{l=1}^{L}t_l$, $h = \prod_{l=1}^{L}h_l$ and $w = \prod_{l=1}^{L}w_l$. We test both linear and 3D convolution layers for the \emph{embedding} layer and observe that using a 3D convolution-based \emph{patchify} module in the tokenizer significantly improves reconstruction performance.
\vspace{-4mm}
\paragraph{Spatial and Temporal Attention}
The output tokens of the \emph{patchify} module at each encoder block are then fed to spatial and temporal attention modules. Given a token volume of size $b \times T_l \times H_l \times W_l \times c_l$ at level $l$,  where $b$ denotes the batch size, spatial reasoning is performed by reshaping the tokens to $(b\cdot T_l) \times (H_l \cdot W_l) \times c_l$ and passing the resulting sequence into a spatial attention module. This is followed by causal attention in the temporal dimension, where the tokens are rearranged into a sequence of size $(b\cdot H_l \cdot W_l) \times T_l \times c_l$. The causal design enables our video model to tokenize single images as well.

Previous sequence-based discrete tokenizers~\cite{villegas2022phenaki,zhao2024image,wang2024omnitokenizer} employ a stack of Transformer~\cite{vaswani2017attention} layers for the spatial and temporal attention modules. However, using Transformers in our hierarchical approach introduced two key challenges. First, the use of positional embeddings makes it difficult to tokenize spatial and temporal resolutions that were not encountered during training, as noted in~\cite{yu2023language}. Our experimental analysis reveals that using embedding extrapolation techniques, such as RoPE~\citep{su2024roformer} or AliBi~\citep{press2021train}, does not fully resolve this issue. Second, the quadratic-scale attention of Transformers imposes a significant computational cost, particularly in the early encoder blocks, rendering high-resolution training and inference impractical. 

To address these challenges, we introduce a Mamba-based tokenizer by incorporating Mamba~\cite{gu2023mamba,dao2024transformers} layers into the spatial and temporal attention modules. This approach is intuitive because Mamba is a powerful model for reasoning over long-sequence inputs and does not require explicit positional encoding, as it operates in a recurrent manner. Consequently, it effectively mitigates the positional encoding bias that limits the generalization capabilities of Transformer-based tokenizers~\cite{villegas2022phenaki}. Furthermore, Mamba’s linear-scale attention significantly enhances computational efficiency compared to Transformer, allowing for high-resolution training and inference.
\vspace{-4mm}
\paragraph{Token Pooling}
To further enhance the hierarchical encoding in our model, we introduce skip connections between the different blocks of the encoder, as depicted in~\Fref{fig:arch_overview}c. Let $v_l$ denote the encoded tokens at level $l$. We downsample the output of the previous encoder block,~\ie~$v_{l-1}$, to match the size of $v_l$, then apply a residual sum before passing it to the encoder block at the next level, $l+1$. While the direct feedforward connections facilitate coarse-to-fine representation learning, the skip connections help retain higher-level features, enabling a more effective encoding of the input video. We use 3D \emph{average pooling} with a kernel size of $t_l \times h_l \times w_l$ for spatio-temporal token pooling. 
\subsection{Decoder}
The decoder network takes the \emph{quantized} representation of the encoded tokens and reconstructs the input video. Mirroring the encoder, the decoder employs hierarchical spatio-temporal upsampling in a bottom-up manner, using a series of decoder blocks. Each block comprises a cascade of causal \emph{temporal attention}, \emph{spatial attention}, and \emph{topixel}, modules, as illustrated in~\Fref{fig:arch_overview}c. Similar to the encoder, we use Mamba~\cite{dao2024transformers} layers for the temporal and spatial attention modules within the decoder.
\vspace{-4mm}
\paragraph{ToPixel} The \emph{topixel} module increases the dimensions of a given token volume both spatially and temporally. It includes an \emph{embedding} layer that uses 3D convolution to project the channel dimension of each token to the desired size, followed by a \emph{pixelshuffle} layer that rearranges the projected tokens into an upsampled spatio-temporal dimension. The \emph{topixel} module at each level $l$ of the decoder mirrors the spatio-temporal kernel, \ie~$t_l \times h_l \times w_l$, of the corresponding \emph{patchify} module for upsampling. 
\vspace{-4mm}
\paragraph{Token Interpolation}We also employ skip connections between the different blocks of the decoder in our model (see~\Fref{fig:arch_overview}c). Let $\hat{v}_l$ represent the decoded tokens at level $l$, where $l=1$ denotes the last block in the decoder. We upsample $\hat{v}_{l+1}$ to match the size of $\hat{v}_l$, then residually add it to $\hat{v}_l$ before passing it to the decoder layers at the next level, $l-1$. For token interpolation, we use \emph{nearest interpolation} in both the spatial and temporal dimensions.
\vspace{-2mm}
\section{Proposed Quantization Scheme}
\vspace{-1mm}
\label{sec:method_quant}
While most prior works use vector quantization (VQ)~\cite{gray1984vector}, recent tokenizers have introduced look-up free quantization schemes, such as LFQ~\cite{yu2023language} and FSQ~\cite{mentzer2023finite}, to address VQ's limitations. Building on these, we propose a more efficient quantization scheme for discrete video tokenization.
\vspace{-1mm}
\subsection{Preliminary: Look-up Free Quantization}
\label{sec:preliminary}
\vspace{-1mm}
Let $V$ denote the input video with size of $T \times H \times W \times 3$. For a spatio-temporal compression rate of $\times thw$, the encoded latent $v$ will have a dimension of $ T/t \times H/h \times W/w \times c$, where $c$ denotes the latent channel size. After the quantization step, the total number of quantized tokens, \ie~the sequence length, will be $\frac{THW}{thw}$.

Given codebook $\calC$ of size $|\calC| = 2^N$, LFQ~\cite{yu2023language} requires $c = N$ and each value of $v$ in the channel dimension is quantized to -1 or 1,~\ie~$\hat{v} = \mathrm{sign}(v) = \mathds{-1}\{v \le 0\} + \mathds{1}\{v > 0\}$, where $\hat{v}$ denotes the quantized latent. This significantly limits the representational expressiveness (power) of LFQ compared to VQ~\cite{gray1984vector}, as the quantized latent fed into the decoder is restricted to binary values, whereas in VQ, the latent values can take any real number,~\ie~$\hat{v} \in \mathbb{R}$. 

 Given the encoded latent $v$, FSQ~\cite{mentzer2023finite} first applies a bounding function $f$, and then rounds to integers. The function $f$ is chosen such that each channel in the quantized latent $\hat{v} = \mathrm{round}(f(v))$ takes one of $L$ \emph{unique} values (\eg~$f: v \rightarrow \lfloor {L/2} \rfloor \tanh(v)$). Due to this condition, FSQ requires a much smaller latent channel size $c = M (< N)$ for a codebook size of $2^N$, where $\prod_{i=1}^{M}L_i = 2^{N}$. For instance, with $|\calC| = 2^{16}$, the channel size for LFQ is $c_{\mathrm{lfq}} = 16$, whereas for FSQ it is $c_{\mathrm{fsq}} = 6$. While the quantized latent in FSQ have more diverse (non-binary) values, the smaller number of channels still limits its overall representational capacity.
 \vspace{-1mm}
\subsection{Channel--Split Quantization}
\vspace{-1mm}
We introduce channel-split quantization to effectively increase the representational power of the quantized latent while maintaining the number of tokens. Our key idea is to exploit the trade-off between the compression and quantization steps during tokenization. First, we increase the channel size of the encoded latent by a factor of $K$,~\ie~ the channel dimension of $v$ will be $c\cdot K$. Then, we \emph{split} the encoded latent in the channel dimension into $K$ groups,~\ie~$v= \{v_1, \ldots, v_K\}$. Finally, each split is independently quantized using either LFQ~\cite{yu2023language} or FSQ~\cite{mentzer2023finite}, which we refer to as channel-split LFQ (\emph{CS-LFQ}) and channel-split FSQ (\emph{CS-FSQ}), respectively. The quantized latents are then concatenated channel-wise before being fed to the decoder, \ie~$\hat{v} = \mathrm{concat}(\hat{v}_1, \ldots, \hat{v}_K)$. To maintain a fair comparison with naive LFQ/FSQ, we compensate for the increased channel size by scaling the encoder’s spatio-temporal compression rate by a factor of $K$,~\ie$\times (thw \cdot K)$. Thus, the sequence length after CS-LFQ/CS-FSQ becomes $\frac{(HWT)}{(thw \cdot K)} \times K = \frac{THW}{thw}$, matching that of LFQ/FSQ as described in \Sref{sec:preliminary}.
\begin{table*}[!t]
    \centering
    \caption{\textbf{Experimental comparison with state-of-the-art} video tokenization and quantization approaches on the video reconstruction task. The best results are highlighted in \textbf{bold}, and the second-best results are \underline{underlined}.}
    \vspace{-2mm}
    \mytabular{0.75}{
    \begin{tabular}{lcccccccccc}
    \toprule
    Method & \makecell{Compression \\ Rate}  & \makecell{Codebook \\ Size} & \makecell{Channel \\ Size} & \makecell{Total \# \\ of Tokens} & \multicolumn{3}{c}{\textbf{Xiph-2K}} & \multicolumn{3}{c}{\textbf{DAVIS}} \\ \cmidrule(lr){6-8} \cmidrule(lr){9-11}
    & $t \times h \times w$ & $|\calC|$ & $c$ & &  PSNR $\uparrow$ & SSIM $\uparrow$ & LPIPS $\downarrow$  & PSNR $\uparrow$ & SSIM $\uparrow$ & LPIPS $\downarrow$ \\ \midrule
    VideoGPT~\cite{yan2021videogpt} (VQ) & $4 \times 4 \times 4$ & $2^{11}$ & $256$ & $THW/64$ & 31.09 & \textbf{0.819} & 0.327 & \underline{31.30} & \textbf{0.771} & 0.305 \\
    CViViT~\cite{villegas2022phenaki} (VQ) & $2 \times 8 \times 8$ & $2^{13}$ & $32$ & $THW/128$ & 28.92 & 0.708 & 0.232 & 27.73 & 0.660 & 0.272 \\
    OmniTokenizer~\cite{wang2024omnitokenizer} (VQ) & $4 \times 8 \times 8$ & $2^{13}$ & $8$ & $THW/256$ & 25.96 & 0.691 & 0.181 & 25.34 & 0.633 & \underline{0.208} \\ \midrule
    Magvit-v2~\cite{yu2023language} (LFQ) & $4 \times 8 \times 8$ & $2^{16}$ & $16$ & $THW/256$ & 30.02 & 0.701 & 0.189 & 29.26 & 0.652 & 0.241\\
    \rowcolor{verylightgray}
    Magvit-v2 + CS-LFQ & $8 \times 8 \times 8$ & $2^{16}$ & $32$ & $THW/256$ & 30.97 & 0.719 & 0.172 & 30.57 & 0.670 & 0.226 \\
    Magvit-v2 + FSQ & $4 \times 8 \times 8$ & $2^{16}$ & $6$ & $THW/256$ & 30.69 & 0.714 & 0.185 & 30.06 & 0.666 & 0.238\\
    \rowcolor{verylightgray}
    Magvit-v2 + CS-FSQ & $8 \times 8 \times 8$ & $2^{16}$ & $12$ & $THW/256$ & 31.08 & 0.728 & 0.165 & 30.75 & 0.681 & 0.214\\
    \midrule
    Ours + LFQ & $4 \times 8 \times 8$ & $2^{16}$ & $16$ & $THW/256$ & 31.05 & 0.711 & 0.171 & 30.02 & 0.669 & 0.224 \\
    \rowcolor{verylightgray}
    Ours + CS-LFQ & $8 \times 8 \times 8$ & $2^{16}$ & $32$ & $THW/256$ & \underline{31.95} & 0.738 & \underline{0.160} & 31.25 & 0.673 & 0.218\\
    Ours + FSQ & $4 \times 8 \times 8$ & $2^{16}$ & $6$ & $THW/256$ & 31.43 & 0.722 & 0.168 & 30.65 & 0.678 & 0.217 \\
    \rowcolor{verylightgray}
    Ours + CS-FSQ & $8 \times 8 \times 8$ & $2^{16}$ & $12$ & $THW/256$ & \textbf{32.54} & \underline{0.747} & \textbf{0.151} & \textbf{32.36} & \underline{0.691} & \textbf{0.206}\\
    \bottomrule
    \end{tabular}
    }
    \vspace{-4mm}
    \label{tab:recon_results}
\end{table*}
\subsection{Why Channel-Split Quantization Works?}
\label{sec:wh_cs_works}
\paragraph{Proposition:} \emph{Given a codebook size of $2^N$, channel-split quantization is an effective technique for increasing representation capacity (achieving an effective single codebook size $\gg 2^N$) while maintaining the total number of tokens.}
\vspace{-4mm}
\paragraph{Proof:} Let the given codebook budget be $|C|=2^N$. In LFQ/FSQ, each pixel in the quantized latent $\hat{v}$ is represented by a \emph{single} token. In contrast, in CS-LFQ/CS-FSQ, each pixel in $\hat{v}$ is represented by a \emph{sequence of $K$} tokens, as the encoded latent is split into $K$ groups in the channel dimension. The \emph{order} of these $K$ tokens for each pixel is important, as each is quantized independently. Let $\{q_1, \ldots, q_K\}$ represent the sequence of $K$ tokens for each pixel in $\hat{v}$. If we were to create a single codebook for CS-LFQ/CS-FSQ, \ie~\textit{represent each pixel in $\hat{v}$ with a single token}, the best \emph{non-overlapping} way to map $\{q_1, \ldots, q_K\}$ into one token would be to use the basis $\{2^{N (K-1)}, 2^{N (K-2)}, \ldots, 2^{N(0)}\}$ and the mapping would be:
\vspace{-1mm}
\begin{equation*}
    f:(q_1, \ldots, q_K) \rightarrow q_1\cdot 2^{N (K-1)} + q_2\cdot 2^{N (K-2)} + \ldots + q_K 
    \vspace{-2mm}
\end{equation*}
With such mapping, the possible maximum token id will be $2 ^N \cdot 2^{N (K-1)} + 2^N \cdot 2^{N (K-2)} + \ldots + 2^N > 2^{NK}$ as the maximum value for each $q_i$ (where $i\in[1,K]$) is $2^N$. Therefore, \textit{CS-LFQ/CS-FSQ, in essence, operates as if using a single codebook of size greater than $ 2^{NK}$}. 
\vspace{-3mm}
\paragraph{Compression Rate vs. Quantization} In LFQ/FSQ-based tokenization, the compression factor is $\times thw$, resulting in more pixels in the encoded latent. In contrast, CS-LFQ/CS-FSQ applies a compression factor of $\times (thw \cdot K)$, leading to fewer pixels. However, CS-LFQ/CS-FSQ offers significantly greater representational power, as it operates with an effective single codebook size of $> 2^{NK}$, compared to LFQ/FSQ's codebook size of $2^N$. Our experimental results confirm that this enhanced representational power compensates for the reduced pixel count, explaining the superior performance of CS-LFQ/CS-FSQ in both reconstruction and generation tasks (refer to \Sref{sec:exp_recon} and \Sref{sec:exp_gen}).
\vspace{-1mm}
\section{Experiment}
\label{sec:exp}
\vspace{-1mm}
\paragraph{Network Training} Following prior works~\cite{yu2021vector,yan2021videogpt}, we train our tokenizer using a standard combination of loss functions: \emph{reconstruction}, \emph{perceptual}, and \emph{GAN} losses. For the reconstruction loss, we minimize the $\calL_1$ distance between the input video 
$V$ and the decoded video $\hat{V}$. For the perceptual loss, we compute the frame-wise LPIPS~\cite{zhang2018unreasonable} between the frames of the input and reconstructed videos. For the GAN loss, we employ a 3D convolution-based PatchGAN discriminator~\cite{isola2017image} to distinguish between real videos and those generated by our model. For FSQ-based tokenizers~\cite{mentzer2023finite}, no loss is applied for codebook training. In the case of LFQ-based tokenizers~\cite{yu2023language}, we follow Yu~\etal~\cite{yu2023language} and incorporate both \emph{entropy penalty} and \emph{commitment} losses during training.
\subsection{Video Tokenization}
\label{sec:exp_recon}
\paragraph{Implementation Details} For LFQ-based tokenizers, the entropy and commitment loss coefficients are set to 0.1 and 0.25, respectively. For FSQ-based models, the levels are set to $[8,8,8,5,5,5]$. In channel-split quantization, we experiment with the number of splits $K$ set to $\{1,2,4\}$. The number of encoder/decoder levels is set to $l = 3$. For $8 \times 8 \times 8$ hierarchical encoding and decoding, the spatio-temporal kernels in the \emph{patchify} and \emph{topixel} modules are set to $[2 \times 4 \times 4]_{l=1}$, $[2 \times 2 \times 2]_{l=2}$, and $[2 \times 1 \times 1]_{l=3}$. Each Mamba layer has a hidden dimension of $512$. We use the WebVid-2M~\cite{Bain21} dataset for model training. At each training step, we randomly sample a video clip of size $16 \times 240 \times 240 \times 3$ with a frame stride of $1$ and feed it into the tokenizer. Each tokenizer is trained for $400$K iterations using the Adam~\citep{kingma2014adam} optimizer with a learning rate of $1\mathrm{e}-4$. The GAN loss is activated at the 200K iteration. We use a batch size of 32 and train across 32 NVIDIA A100 GPUs.

\vspace{-3mm}
\paragraph{Baseline Methods} 
We benchmark our approach against several discrete video tokenizers with publicly available code. These include VideoGPT~\cite{yan2021videogpt}, CViViT~\cite{villegas2022phenaki}, and OmniTokenizer~\cite{wang2024omnitokenizer}, all of which use VQ, as well as the current state-of-the-art model, Magvit-v2~\cite{yu2023language}, which employs LFQ. Additionally, we compare various quantization schemes, including LFQ, FSQ, and channel-split quantization (CS-LFQ and CS-FSQ), using both Magvit-v2 (causal 3D convolution-based) and our Mamba-based tokenizer. All models are trained under the same settings as ours, strictly adhering to their official implementations.
\vspace{-3mm}
\paragraph{Evaluation Dataset and Metrics}
We evaluate our model and competing approaches on the video reconstruction task using two representative datasets with medium to large motion: Xiph-2K~\cite{Niklaus_CVPR_2020} and DAVIS~\cite{pont20172017}. During evaluation, we use a frame sequence length of 16 at a resolution of 480p. The quality of the reconstructed video is assessed using PSNR, SSIM, and LPIPS~\cite{zhang2018unreasonable} metrics.

\begin{figure*}[!t]
    \centering
    \includegraphics[width=1\linewidth, trim={1.65cm, 8.0cm, 3.55cm, 4.2cm}, clip]{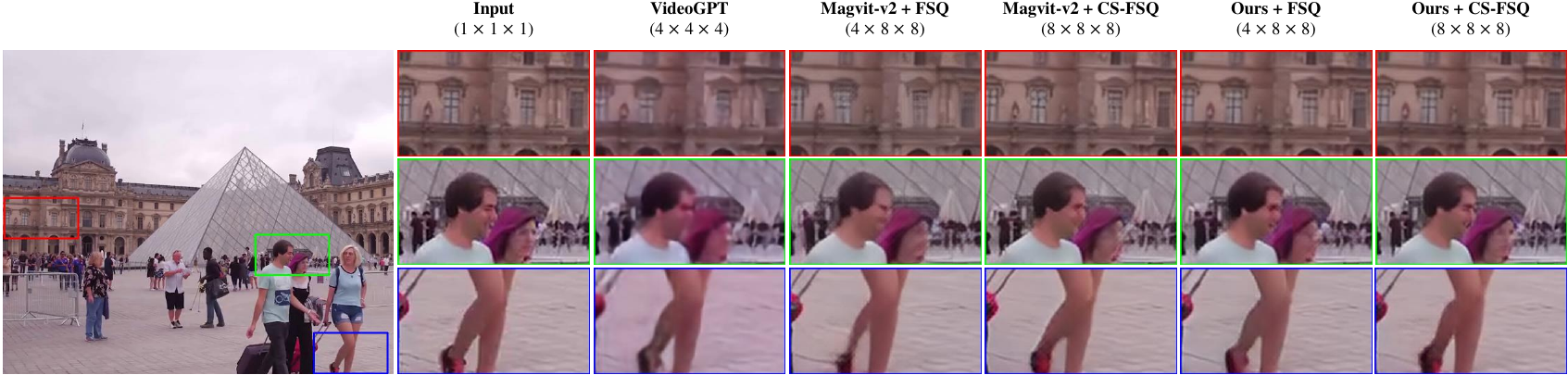}
    \caption{\textbf{Qualitative analysis} of our tokenizer compared with the best-performing baselines on the video reconstruction task.}
    \label{fig:qual_compare}
    \vspace{-5mm}
\end{figure*}
\subsubsection{Results}
\label{sec:recon_results}
In \Tref{tab:recon_results}, we provide a comprehensive comparison of our approach with state-of-the-art video tokenizers~\cite{yan2021videogpt,villegas2022phenaki,wang2024omnitokenizer,yu2023language} and quantization methods~\cite{gray1984vector,yu2023language,mentzer2023finite} for the video reconstruction task. As shown in~\Tref{tab:recon_results}, our Mamba-based tokenizer demonstrates strong performance, consistently surpassing the current state-of-the-art, Magvit-v2~\cite{yu2023language}, across various quantization schemes. For example, our tokenizer with LFQ quantization (\emph{Ours + LFQ}) achieves an average performance of 30.54 dB on the Xiph-2K and DAVIS datasets, outperforming Magvit-v2, which attains only 29.64 dB. Moreover, our best configuration (\emph{Ours + CS-FSQ}) exceeds Magvit-v2 by 2.81 dB and CViViT by 4.1 dB, even with twice the temporal compression. This improvement stems from key architectural choices, including hierarchical downsampling and upsampling via 3D convolution-based \emph{patchify} and \emph{topixel} layers, skip connections through token pooling and interpolation, and efficient spatio-temporal attention using Mamba layers.

\Tref{tab:recon_results} further demonstrates that the proposed channel-split quantization significantly improves the performance of LFQ and FSQ across both causal 3D convolution-based (Magvit-v2) and Mamba-based (ours) models, while maintaining the number of tokens. For example, \emph{Magvit-v2 + LFQ} achieves an average reconstruction performance of 29.64 dB across the two datasets, whereas \emph{Magvit-v2 + CS-LFQ} improves to 30.77 dB (+1.13 dB). Likewise, \emph{Ours + CS-FSQ} surpasses \emph{Ours + FSQ} by an average margin of 1.41 dB. This notable performance gain is mainly due to the enhanced representational capacity of the quantized latent space enabled by channel-split quantization, as elaborated in~\Sref{sec:wh_cs_works}. This enhancement facilitates better video decoding, even at a high compression rate ($\times 512$). 

In \Fref{fig:qual_compare}, we present a qualitative comparison of our approach against the top-performing baselines from \Tref{tab:recon_results}. As shown in the figure, \emph{Magvit-v2 + FSQ} struggles to faithfully decode fast motions (see the \emph{legs} in the \textcolor{blue}{blue} box), preserve facial details (see the \emph{faces} in the \textcolor{green}{green} box), and maintain the structural details of distant objects (see the \emph{windows} in the \textcolor{red}{red} box). In comparison, our Mamba-based model (\emph{Ours + FSQ}) reconstructs frames with enhanced sharpness and detail. Moreover, Magvit-v2 with channel-split quantization (\emph{Magvit-v2 + CS-FSQ}) demonstrates a significant improvement in decoded frame quality over its base model (\emph{Magvit-v2 +FSQ}). Our best model (\emph{Ours + CS-FSQ}) not only preserves the structural details of small objects located far from the camera, but also reconstructs facial features with high fidelity, even at a $\times 512$ compression rate, as depicted in \Fref{fig:qual_compare}.

\vspace{-1mm}
\subsection{Video Generation}
\label{sec:exp_gen}
\vspace{-1mm}
One of the primary applications of our work is video generation, where the encoder compresses input video into quantized tokens for generative modeling, and the decoder reconstructs a video from generated tokens. To demonstrate this, we integrate our pretrained video tokenizer, along with tokenizers from competing approaches~\cite{yan2021videogpt,yu2023language}, into the open-source autoregressive framework VideoGPT~\cite{yan2021videogpt} and train each for unconditional video generation.
\begin{table}[!t]
    \centering
    \caption{\textbf{Experimental results} on unconditional video generation.}
    \vspace{-2mm}
    \mytabular{0.72}{
    \begin{tabular}{lccc}
    \toprule
    \textbf{Tokenizer} + \textbf{Generator} & \makecell{Comp. \\ Rate} & \textbf{SkyTimelapse} & \textbf{UCF-101} \\ \cmidrule(lr){3-3} \cmidrule(lr){4-4}
     & $t \times h \times w$ & $\mathrm{FVD}$ $\downarrow$ & $\mathrm{FVD}$ $\downarrow$ \\ \midrule
    VideoGPT (VQ) + VideoGPT & $4 \times 4 \times 4$ & 129.2 &  423.1 \\ \midrule
    Magvit-v2 (LFQ) + VideoGPT  & $4 \times 8 \times 8$ & 96.2 & 376.7 \\
    \rowcolor{verylightgray}
    Magvit-v2 (CS-LFQ) + VideoGPT & $8 \times 8 \times 8$ & 81.7 & 330.5\\
    Magvit-v2 (FSQ) + VideoGPT  & $4 \times 8 \times 8$ & 84.0 & 339.3 \\
    \rowcolor{verylightgray}
    Magvit-v2 (CS-FSQ) + VideoGPT & $8 \times 8 \times 8$ & 71.4 & 289.6\\ \midrule
    Ours (LFQ) + VideoGPT & $4 \times 8 \times 8$ & 79.9 & 323.3 \\
    \rowcolor{verylightgray}
    Ours (CS-LFQ) + VideoGPT  & $8 \times 8 \times 8$ & \underline{62.2} & \underline{280.7} \\
    Ours (FSQ) + VideoGPT  & $4 \times 8 \times 8$ & 70.1 &  293.4 \\
    \rowcolor{verylightgray}
    Ours (CS-FSQ) + VideoGPT & $8 \times 8 \times 8$ & \textbf{55.4} &  \textbf{266.2} \\ \bottomrule
    \end{tabular}
    }
    \vspace{-4mm}
    \label{tab:gen_results}
\end{table}
\paragraph{Implementation Details}
We use the training split of commonly used video synthesis benchmarks, SkyTimelapse~\cite{dtvnet} and UCF-101~\citep{soomro2012ucf101}, for our video generation experiments. During training, each frame in the datasets is resized to a resolution of $256\times 256$, and we sample video clips consisting of $16$ frames with a frame stride of $1$. Our experiments are conducted on 32 NVIDIA A100 GPUs, following the training configuration utilized in VideoGPT~\cite{yan2021videogpt}.
\vspace{-3mm}
\paragraph{Baseline Methods} 
We establish multiple baselines by integrating various pretrained tokenizers, as listed in \Tref{tab:gen_results}, into the VideoGPT framework. These include the Magvit-v2 tokenizer and our Mamba-based tokenizer, each employing different quantization methods such as LFQ, FSQ, and channel-split quantization (CS-LFQ and CS-FSQ). We benchmark these baselines against VideoGPT’s original VQ-based tokenizer.
\vspace{-3mm}
\paragraph{Evaluation Metrics} To assess the quality of the generated video clips, we use the Fréchet Video Distance (FVD)~\cite{unterthiner2018towards} metric. Following the evaluation protocols of prior works~\cite{PVDM, skorokhodov2022stylegan}, we calculate the FVD score on 2,048 real and generated video clips, each consisting of 16 frames.

\subsubsection{Results}
In \Tref{tab:gen_results}, we present a quantitative evaluation of the videos generated by VideoGPT using various pretrained tokenizer models. As shown in the table, VideoGPT's VQ-based tokenizer performs worse than both Magvit-v2 and our tokenizer, likely due to the challenges of long-sequence modeling arising from VideoGPT's lower spatio-temporal compression rate. In contrast, both Magvit-v2 and our tokenizer operate on sequences that are $4 \times$ shorter, resulting in significantly improved performance. Notably, VideoGPT enabled by our Mamba-based tokenizer,~\ie~\emph{Ours (CS-FSQ) + VideoGPT}, achieves the best generation performance, surpassing its Magvit-v2 counterpart,~\ie~\emph{Magvit-v2 (CS-FSQ) + VideoGPT}, by a notable margin. This result underscores the proposed tokenizer's potential for autoregressive video generation. Furthermore, as shown in \Tref{tab:gen_results}, channel-split quantization-based tokenizers (\ie~CS-LFQ and CS-FSQ) consistently outperform their base counterparts (\ie~LFQ and FSQ) across different models. These findings highlight that channel-split quantization not only enhances the representational power of quantized latents but also produces video tokens that are better suited for generative modeling. 

\begin{figure}[!t]
\begin{center}
\setlength{\tabcolsep}{0.1pt}
\renewcommand{\arraystretch}{0.1}
\resizebox{1.0\linewidth}{!}{%
\begin{tabular}{ccccccccc}
\raisebox{1\normalbaselineskip}{\rotatebox[origin=c]{90}{\scalebox{0.65}{\tiny SkyTimelapse}}} &
\includegraphics[width=0.1\linewidth]{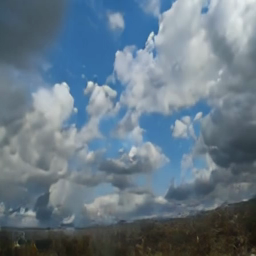}&
\includegraphics[width=0.1\linewidth]{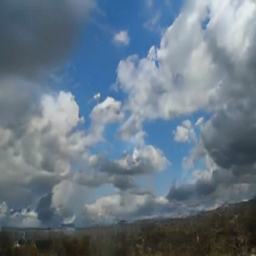}&
\includegraphics[width=0.1\linewidth]{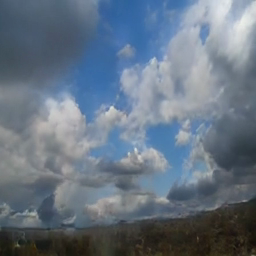}&
\includegraphics[width=0.1\linewidth]{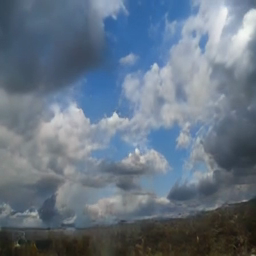}&
\includegraphics[width=0.1\linewidth]{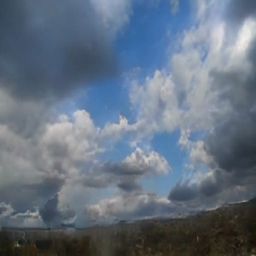}&
\includegraphics[width=0.1\linewidth]{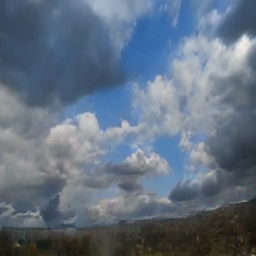}\\
\raisebox{1\normalbaselineskip}{\rotatebox[origin=c]{90}{\scalebox{0.65}{\tiny SkyTimelapse}}} &
\includegraphics[width=0.1\linewidth]{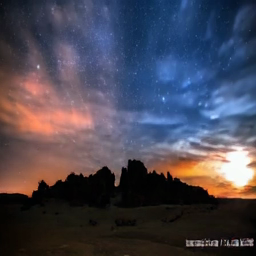}&
\includegraphics[width=0.1\linewidth]{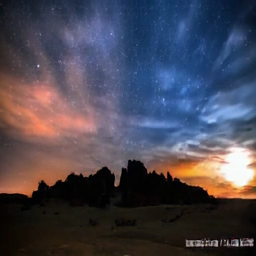}&
\includegraphics[width=0.1\linewidth]{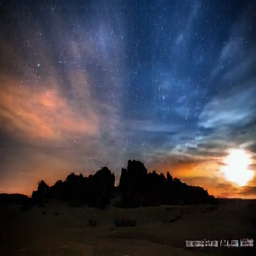}&
\includegraphics[width=0.1\linewidth]{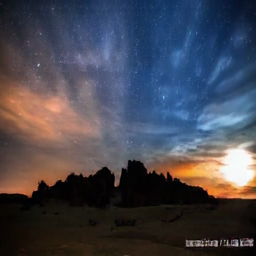}&
\includegraphics[width=0.1\linewidth]{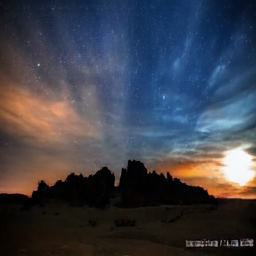}&
\includegraphics[width=0.1\linewidth]
{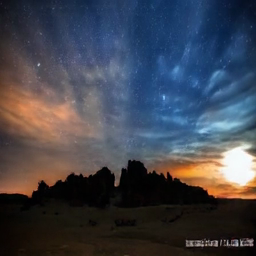}\\
\raisebox{1\normalbaselineskip}{\rotatebox[origin=c]{90}{\scalebox{0.65}{\tiny UCF-101}}} &
\includegraphics[width=0.1\linewidth]{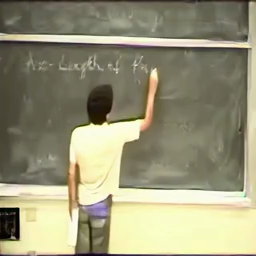}&
\includegraphics[width=0.1\linewidth]{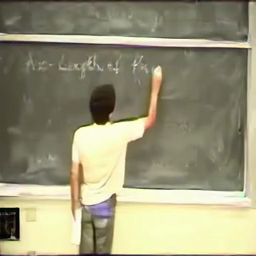}&
\includegraphics[width=0.1\linewidth]{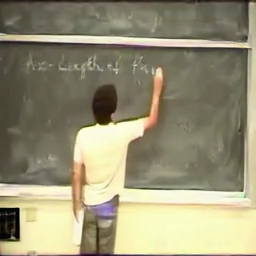}&
\includegraphics[width=0.1\linewidth]{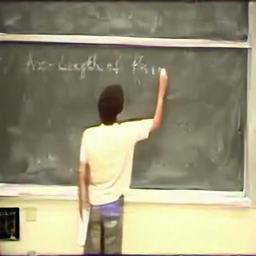}&
\includegraphics[width=0.1\linewidth]{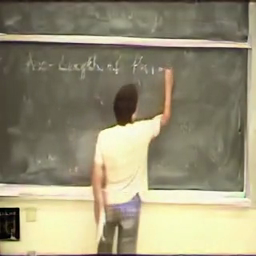}&
\includegraphics[width=0.1\linewidth]{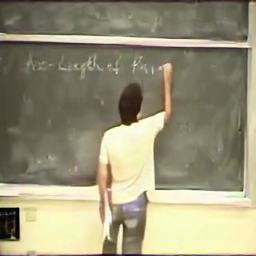}\\
\raisebox{1\normalbaselineskip}{\rotatebox[origin=c]{90}{\scalebox{0.65}{\tiny UCF-101}}} &
\includegraphics[width=0.1\linewidth]{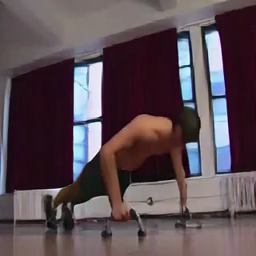}&
\includegraphics[width=0.1\linewidth]{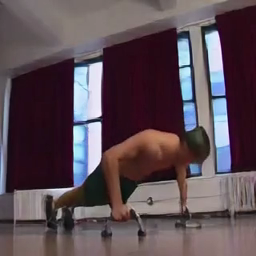}&
\includegraphics[width=0.1\linewidth]{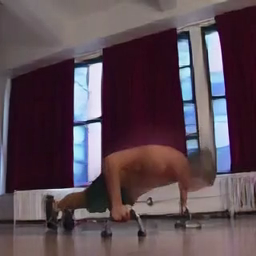}&
\includegraphics[width=0.1\linewidth]{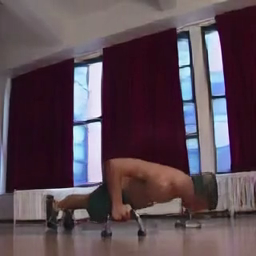}&
\includegraphics[width=0.1\linewidth]{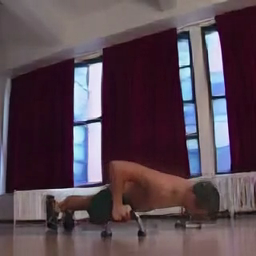}&
\includegraphics[width=0.1\linewidth]{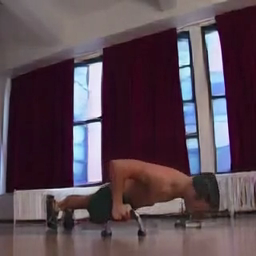}
\end{tabular}}
\end{center}
\vspace{-6.5mm}
\caption{\textbf{Qualitative analysis} of videos generated by VideoGPT, enabled by our tokenizer,~\ie~\emph{Ours (CS-FSQ) + VideoGPT.}}
\label{fig:qual_gen}
\vspace{-5mm}
\end{figure}

In \Fref{fig:qual_gen}, we visualize sequences of video frames generated by VideoGPT empowered with our tokenizer,~\ie~\emph{Ours (CS-FSQ) + VideoGPT}. The first two rows show results from the SkyTimelapse dataset, while the last two correspond to the UCF-101 dataset. As illustrated in the figure, VideoGPT, equipped with our tokenizer, generates realistic sky time-lapse videos and synthesizes human action videos with strong spatio-temporal consistency.

\vspace{-1mm}
\section{Ablation Studies}
\label{sec:ablation}
\vspace{-1mm}
We conduct ablation experiments to analyze the contributions of various design choices in our proposed video tokenizer. All ablations are conducted on a tokenizer with a spatio-temporal compression rate of $8 \times 8 \times 8$, using CS-FSQ with 2 splits ($K=2$). The results on the Xiph-2K~\cite{Niklaus_CVPR_2020} and DAVIS~\cite{pont20172017} datasets are summarized in \Tref{tab:ablation}.
\begin{figure*}[!t]
    \centering
    \includegraphics[width=1\linewidth, trim={3.2cm, 9.9cm, 3.75cm, 4.2cm}, clip]{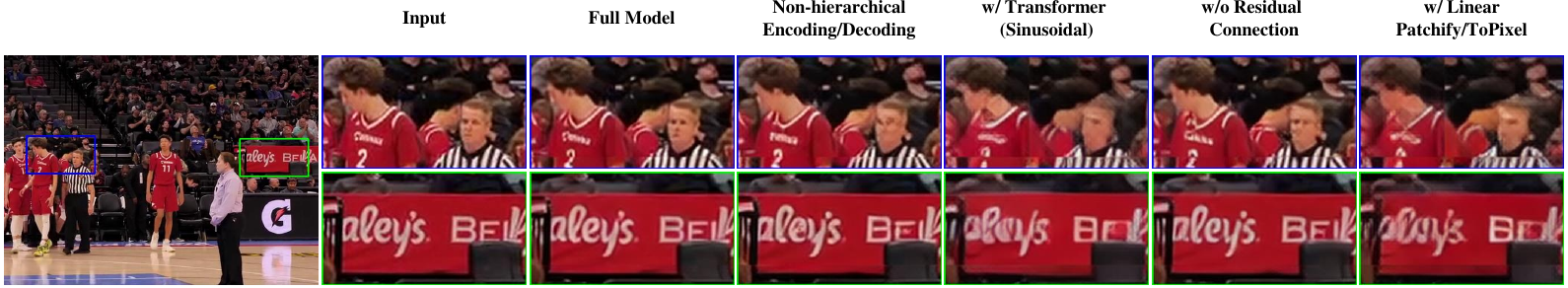}
    \caption{\textbf{Quantitative analysis} of ablation experiments on the video reconstruction task. }
    \label{fig:ablation_qual}
    \vspace{-5mm}
\end{figure*}
\vspace{-3.5mm}
\paragraph{Encoding/Decoding} Here, we study the advantages of hierarchical downsampling and upsampling in discrete video tokenization. To do this, we train a tokenizer model with a single \emph{patchify} layer that directly downsamples the input video using an $8 \times 8 \times 8$ kernel, and a single \emph{topixel} layer that directly upsamples the decoded latent to the input video size with an $8 \times 8 \times 8$ kernel (see \Fref{fig:arch_overview}a). This approach is similar to CViViT~\cite{villegas2022phenaki}, but it replaces Transformers with Mamba layers for attention and substitutes linear layers with 3D convolutions in the embedding layer. In \Tref{tab:ablation}a, we compare this baseline with our model, which incorporates hierarchical spatio-temporal downsampling and upsampling (see \Fref{fig:arch_overview}c). As shown in the table, our hierarchical tokenizer consistently outperforms the non-hierarchical baseline by a significant margin. For example, the non-hierarchical model achieves an average of 30.97 dB across the two datasets, whereas the hierarchical model reaches 32.45 dB, a gain of +1.48 dB. Qualitative results in \Fref{fig:ablation_qual} further illustrate that hierarchical down/upsampling (\emph{Full Model} in \Fref{fig:ablation_qual}) yields sharper decoded frames compared to direct (non-hierarchical) encoding/decoding.
\vspace{-3.5mm}
\begin{table}[!t]
    \centering
    \caption{\textbf{Ablation experiments} on different network components in our video tokenizer.}
    \vspace{-2mm}
    \mytabular{0.7}{
    \begin{tabular}{llcccc}
    \toprule
    & Method & \multicolumn{2}{c}{\textbf{Xiph-2K}} & \multicolumn{2}{c}{\textbf{DAVIS}} \\ \cmidrule(lr){3-4} \cmidrule(lr){5-6}
    & & PSNR $\uparrow$ & LPIPS $\downarrow$  & PSNR $\uparrow$ & LPIPS $\downarrow$ \\ \midrule
    (a) & \textbf{Encoding/Decoding} & & & & \\
     & Non-hierarchical & 31.18 & 0.213 & 30.77 & 0.251  \\ 
     & Hierarchical & 32.54 & 0.151 & 32.36 & 0.206  \\
     \midrule
    (b) & \textbf{Spatial/Temporal Attention}& & & & \\ 
     & Transformer (Sinusoidal) & 30.69 & 0.218 & 30.54 & 0.253 \\
     & Transformer (RoPE) & 30.82 & 0.212 & 30.78 & 0.246\\
     & Transformer (AliBi) & 31.06 & 0.210 & 30.91 & 0.243\\
     & Mamba & 32.54 & 0.151 & 32.36 & 0.206  \\
    \midrule
    (c) & \textbf{Token Pooling/Interpolation} & & & & \\ 
    & Feedforward & 31.70 & 0.205 & 31.68 & 0.237 \\ 
    & Feedforward + Residual & 32.54 & 0.151 & 32.36 & 0.206  \\
    \midrule
    (d) & \textbf{Patchify/ToPixel Modules} & & & & \\
    & Linear & 30.34 & 0.225 & 30.05 & 0.268\\
    & 3D Convolution & 32.54 & 0.151 & 32.36 & 0.206  \\
    \bottomrule
    \end{tabular}
    }
    \vspace{-4mm}
    \label{tab:ablation}
\end{table}
\paragraph{Spatial/Temporal Attention}
We compare different network architectures for the spatial and temporal attention modules in our tokenizer model, including Transformer~\cite{vaswani2017attention} and Mamba~\cite{dao2024transformers}. For the Transformer-based tokenizer, we experiment with three types of positional encoding mechanisms: Sinusoidal~\cite{vaswani2017attention}, RoPE~\citep{su2024roformer}, and AliBi~\citep{press2021train}. As indicated in the \Tref{tab:ablation}b, using Transformer layers leads to subpar tokenization performance. This is primarily because the positional embeddings make it difficult to tokenize spatial resolutions not encountered during training, as noted in Yu~\etal~\cite{yu2023language}. While embedding extrapolation techniques such as RoPE~\citep{su2024roformer} or AliBi~\citep{press2021train} slightly improve performance, they do not fully address the problem, as can be inferred from \Tref{tab:ablation}b. In comparison, our Mamba-based tokenizer achieves strong performance both quantitatively and qualitatively (see \emph{Full Model} in \Fref{fig:ablation_qual}). Given the capability of Mamba layers to effectively reason over long sequences without requiring positional embeddings~\cite{dao2024transformers}, they are an ideal choice for our sequence-based video tokenizer, explaining the superior results observed in \Tref{tab:ablation}b.

\paragraph{Token Pooling/Interpolation} We investigate the benefits of adding residual connections within the encoder blocks (using token pooling) and decoder blocks (using token interpolation) as discussed in \Sref{sec:method_enc_dec}. To do this, we train our tokenizer model without residual connections,~\ie~the encoder and decoder blocks are connected only through feedforward pathways. In \Tref{tab:ablation}c, we compare this baseline with our model, which includes residual (skip) connections. As shown in the table, introducing residual connections via token pooling/interpolation results in an average performance boost of 0.76 dB. Additionally, it can be inferred from \Fref{fig:ablation_qual} that a model incorporating residual connections within the encoder and decoder blocks maintains sharper details in the reconstructed video frames (see \emph{Full Model} in \Fref{fig:ablation_qual}) compared to a model without residual connections.
\vspace{-3.5mm}
\paragraph{Patchify/ToPixel Modules} We compare different architectural choices for the \emph{embedding} layer in the \emph{patchify} and \emph{topixel} modules of the tokenizer model. We experiment with both a linear layer and a 3D convolution layer. As noted in \Tref{tab:ablation}d, our tokenizer using 3D convolution in the \emph{patchify}/\emph{topixel} modules outperforms its linear layer-based counterpart by a significant margin of 2.25 dB on average. This result underscores the importance of the embedding layer, particularly for sequence-based video tokenization, as 3D convolutions are better suited to handle spatio-temporal dependencies across video frames compared to linear layers. A similar conclusion can be drawn from the qualitative analysis in \Fref{fig:ablation_qual}, where a tokenizer with a linear embedding layer struggles to decode facial features and structural details of distant objects. In contrast, our tokenizer with a 3D convolution-based embedding layer (\emph{Full Model} in \Fref{fig:ablation_qual}) faithfully reconstructs high-quality frames.

\vspace{-1mm}
\section{Experimental Analysis}
\label{sec:cs_appx2}
\vspace{-1mm}
Here, we investigate the compression-quantization trade-off in video tokenization using channel-split quantization. Specifically, we examine video reconstruction performance at higher spatio-temporal compression rates under different channel-split quantization configurations. We use Magvit-v2~\cite{yu2023language} as the base model and FSQ~\cite{mentzer2023finite} as the primary quantization method. As shown in \Tref{tab:higher_comp}, we experiment with \emph{effective} compression rates of $\times 512$ and $\times 1024$, ~\ie~$\frac{thw}{K}$, where $K=1$ represents FSQ and $K > 1$ represents CS-FSQ. Notably, for smaller splits (\eg, $K=2$), channel-split FSQ significantly outperforms naive FSQ at both $\times 512$ and $\times 1024$ compression rates. For example, CS-FSQ ($K=2$) at a $16 \times 8 \times 8$ compression rate achieves notably better performance than FSQ ($K=1$) at an $8 \times 8 \times 8$ compression rate. A similar trend is observed when comparing CS-FSQ ($K=2$) with an $8 \times 16 \times 16$ compression rate to FSQ ($K=1$) with a $4 \times 16 \times 16$ compression rate.

However, our experiments reveal that the performance gains from channel-split quantization plateau as the number of splits increases (\ie~as the spatio-temporal compression rate increases). As shown in \Tref{tab:higher_comp}, CS-FSQ ($K=4$) at an $8 \times 16 \times 16$ compression rate performs on par with FSQ ($K=1$) at an $8 \times 8 \times 8$ compression rate. Similarly, CS-FSQ ($K=4$) at a $16 \times 16 \times 16$ compression rate offers no improvement over FSQ ($K=1$) at a $4 \times 16 \times 16$ compression rate. We hypothesize that \emph{at very high compression rates, the compression-quantization trade-off becomes dominated by compression}. As a result, increasing the representational capacity of the latent encoding through channel-split quantization provides little benefit for video tokenization, as the extreme dimensionality reduction limits its effectiveness.

\begin{table}[!t]
    \centering
    \caption{\textbf{Experimental analysis} on the quantization-compression trade-off on channel split quantization.}
    \vspace{-2mm}
    \mytabular{0.72}{
    \begin{tabular}{lllcccc}
    \toprule
     \makecell[l]{Compression \\ Rate}  & \makecell[l]{Channel \\ Size} & \makecell[l]{\# of \\ Splits} & \multicolumn{2}{c}{\textbf{Xiph-2K}} & \multicolumn{2}{c}{\textbf{DAVIS}} \\ \cmidrule(lr){4-5} \cmidrule(lr){6-7}
    $t \times h \times w$ & $c$ & $K$ & PSNR $\uparrow$ & LPIPS $\downarrow$  & PSNR $\uparrow$ & LPIPS $\downarrow$ \\ \midrule
    $8 \times 8 \times 8$ & 6 & 1 & 29.00 & 0.215 & 28.34 & 0.278 \\
    \rowcolor{verylightgray}
    $16 \times 8 \times 8$ & 12 & 2 & 29.74 & 0.202 & 29.02 & 0.256\\
    \rowcolor{verylightgray}
    $4 \times 16 \times 16$ & 12 & 2 & 29.44 & 0.210 & 28.76 & 0.272 \\
    $8 \times 16 \times 16$ & 24 & 4 & 29.02 & 0.220 & 28.30 & 0.288 \\ \midrule
    $4 \times 16 \times 16$ & 6 & 1 & 26.12 & 0.308 & 25.53 & 0.327 \\
    \rowcolor{verylightgray}
    $8 \times 16 \times 16$ & 12 & 2 & 26.88 & 0.286 & 26.17 & 0.308 \\
    $16 \times 16 \times 16$ & 24 & 4 & 26.10 & 0.310 & 25.64 & 0.330\\
    \bottomrule
    \end{tabular}
    }
    \label{tab:higher_comp}
    \vspace{-6mm}
\end{table}
\vspace{-2mm}
\section{Conclusion}
\vspace{-1mm}
Our work makes two key contributions to discrete video tokenization. First, we propose a Mamba-based encoder-decoder architecture that overcomes the limitations of previous tokenizers. Second, we introduce channel-split quantization to enhance the representational power of quantized latents without increasing token count. Our model establishes a new state-of-the-art in both video tokenization and generation, outperforming both causal 3D convolution and Transformer-based approaches across multiple datasets.

\section{Appendix}
Here, we present additional experimental analysis that complements our findings in the main paper. 
\begin{figure*}[!htb]
    \centering
    \includegraphics[width=1\linewidth, trim={1.65cm, 5.95cm, 1.7cm, 3.7cm}, clip]{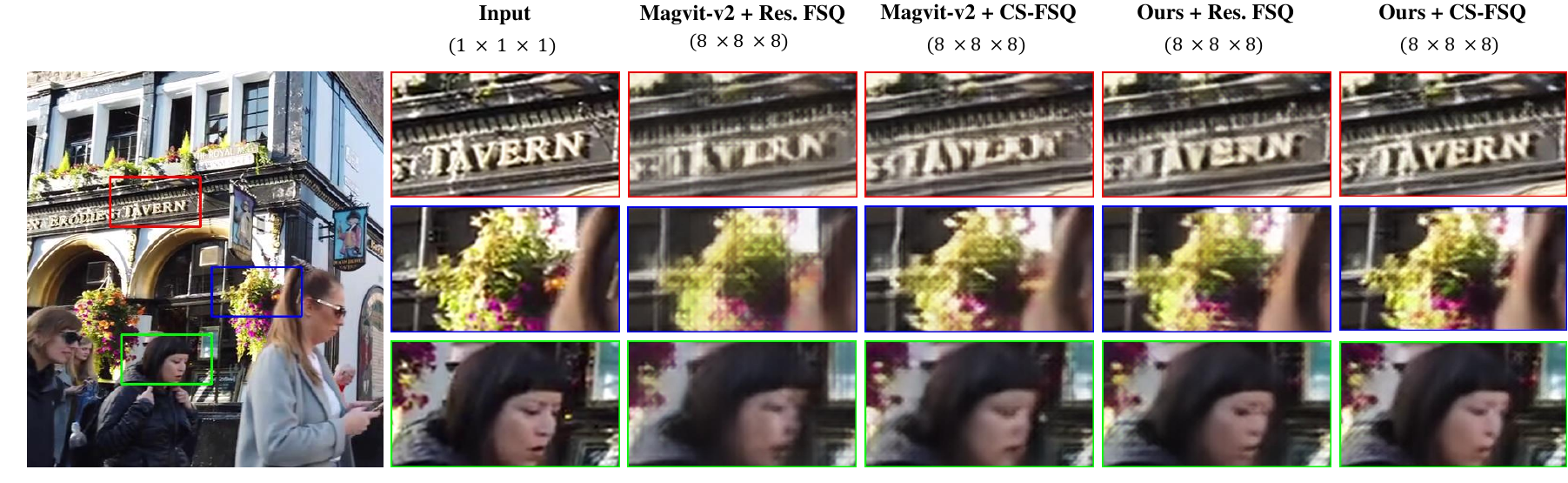}
    \vspace{-6mm}
    \caption{\textbf{Qualitative analysis} of residual and channel-split with Magvit-v2 and our proposed tokenizer.}
    \label{fig:qual_compare_supp}
\end{figure*}
\begin{table}[!htb]
    \centering
    \caption{\textbf{Experimental comparison} between channel-split and residual quantization}
    \vspace{-2mm}
    \mytabular{0.8}{
    \begin{tabular}{lcccc}
    \toprule
    Method & \multicolumn{2}{c}{\textbf{Xiph-2K}} & \multicolumn{2}{c}{\textbf{DAVIS}} \\ \cmidrule(lr){2-3} \cmidrule(lr){4-5}
    & PSNR $\uparrow$ & LPIPS $\downarrow$  & PSNR $\uparrow$  & LPIPS $\downarrow$ \\ \midrule
    Magvit-v2 + Res. FSQ & 30.60 & 0.187 & 30.08 & 0.268 \\
    \rowcolor{verylightgray}
    Magvit-v2 + CS-FSQ & 32.84 & 0.138 & 31.97 & 0.193\\
    Ours + Res. FSQ  & 32.68 & 0.144 & 32.15 & 0.186 \\
    \rowcolor{verylightgray}
    Ours + CS-FSQ & 34.47 & 0.114 & 34.34 & 0.172\\ \bottomrule
    \end{tabular}
    }
    \vspace{-2mm}
    \label{tab:cs_vs_res}
\end{table}
\subsection{Channel-Split vs. Residual Quantization}
\label{sec:cs_appx1}
In this section, we compare the proposed channel-split quantization with residual quantization. In the residual quantization scheme~\cite{lee2022autoregressive,adiban2023s,tian2024visual,defossez2022high}, the encoded latent $v$  is first quantized to $\hat{v}$, after which the residual $v-\hat{v}$ is computed and subsequently quantized. This process is repeated for a predefined number of steps. The quantized representations from each residual step are then combined through summation and fed into the decoder. Note that residual quantization leads to an increase in the number of tokens required for generative modeling. In the case of \emph{residual} LFQ/FSQ, if the number of residual steps is set to $r$, the token sequence length becomes $\frac{HWT}{hwt} \times r$. For channel-split quantization,~\ie~CS-LFQ/CS-FSQ, the same sequence length is achieved by performing spatio-temporal compression at a factor of $\times hwt$ and quantizing across $r$ splits,~\ie,~the encoded latent channel dimension becomes $c \cdot r$. 

In \Tref{tab:cs_vs_res}, we compare \emph{residual FSQ} and \emph{CS-FSQ} across different tokenizer architectures, using $r = 4$ and a spatio-temporal compression rate of $8 \times 8 \times 8$. As shown in the table, channel-split quantization-based tokenizers (\emph{Magvit-v2 + CS-FSQ} and \emph{Ours + CS-FSQ}) demonstrate significantly better performance compared to their residual quantization-based counterparts (\emph{Magvit-v2 + Res. FSQ} and \emph{Ours + Res. FSQ}). The qualitative analysis in \Fref{fig:qual_compare_supp} further illustrates that channel-split quantization consistently yields noticeably sharper frame reconstructions than residual quantization, across both Magvit-v2 and our tokenizer, while preserving the same number of tokens.


\begin{thebibliography}{4}
\providecommand{\natexlab}[1]{#1}
\providecommand{\url}[1]{\texttt{#1}}
\expandafter\ifx\csname urlstyle\endcsname\relax
  \providecommand{\doi}[1]{doi: #1}\else
  \providecommand{\doi}{doi: \begingroup \urlstyle{rm}\Url}\fi

\bibitem[Adiban et~al.(2023)Adiban, Stefanov, Siniscalchi, and
  Salvi]{adiban2023s}
Mohammad Adiban, Kalin Stefanov, Sabato~Marco Siniscalchi, and Giampiero Salvi.
\newblock S-hr-vqvae: Sequential hierarchical residual learning vector
  quantized variational autoencoder for video prediction.
\newblock \emph{arXiv preprint arXiv:2307.06701}, 2023.

\bibitem[D{\'e}fossez et~al.(2022)D{\'e}fossez, Copet, Synnaeve, and
  Adi]{defossez2022high}
Alexandre D{\'e}fossez, Jade Copet, Gabriel Synnaeve, and Yossi Adi.
\newblock High fidelity neural audio compression.
\newblock \emph{arXiv preprint arXiv:2210.13438}, 2022.

\bibitem[Lee et~al.(2022)Lee, Kim, Kim, Cho, and Han]{lee2022autoregressive}
Doyup Lee, Chiheon Kim, Saehoon Kim, Minsu Cho, and Wook-Shin Han.
\newblock Autoregressive image generation using residual quantization.
\newblock In \emph{Proceedings of the IEEE/CVF Conference on Computer Vision
  and Pattern Recognition}, pages 11523--11532, 2022.

\bibitem[Tian et~al.(2024)Tian, Jiang, Yuan, Peng, and Wang]{tian2024visual}
Keyu Tian, Yi Jiang, Zehuan Yuan, Bingyue Peng, and Liwei Wang.
\newblock Visual autoregressive modeling: Scalable image generation via
  next-scale prediction.
\newblock \emph{arXiv preprint arXiv:2404.02905}, 2024.

\end{thebibliography}


{
    \small
    \begin{thebibliography}{33}
\providecommand{\natexlab}[1]{#1}
\providecommand{\url}[1]{\texttt{#1}}
\expandafter\ifx\csname urlstyle\endcsname\relax
  \providecommand{\doi}[1]{doi: #1}\else
  \providecommand{\doi}{doi: \begingroup \urlstyle{rm}\Url}\fi

\bibitem[Adiban et~al.(2023)Adiban, Stefanov, Siniscalchi, and Salvi]{adiban2023s}
Mohammad Adiban, Kalin Stefanov, Sabato~Marco Siniscalchi, and Giampiero Salvi.
\newblock S-hr-vqvae: Sequential hierarchical residual learning vector quantized variational autoencoder for video prediction.
\newblock \emph{arXiv preprint arXiv:2307.06701}, 2023.

\bibitem[Bain et~al.(2021)Bain, Nagrani, Varol, and Zisserman]{Bain21}
Max Bain, Arsha Nagrani, G{\"u}l Varol, and Andrew Zisserman.
\newblock Frozen in time: A joint video and image encoder for end-to-end retrieval.
\newblock In \emph{IEEE International Conference on Computer Vision}, 2021.

\bibitem[Dao and Gu(2024)]{dao2024transformers}
Tri Dao and Albert Gu.
\newblock Transformers are ssms: Generalized models and efficient algorithms through structured state space duality.
\newblock \emph{arXiv preprint arXiv:2405.21060}, 2024.

\bibitem[D{\'e}fossez et~al.(2022)D{\'e}fossez, Copet, Synnaeve, and Adi]{defossez2022high}
Alexandre D{\'e}fossez, Jade Copet, Gabriel Synnaeve, and Yossi Adi.
\newblock High fidelity neural audio compression.
\newblock \emph{arXiv preprint arXiv:2210.13438}, 2022.

\bibitem[Dosovitskiy(2020)]{dosovitskiy2020image}
Alexey Dosovitskiy.
\newblock An image is worth 16x16 words: Transformers for image recognition at scale.
\newblock \emph{arXiv preprint arXiv:2010.11929}, 2020.

\bibitem[Esser et~al.(2021)Esser, Rombach, and Ommer]{esser2021taming}
Patrick Esser, Robin Rombach, and Bjorn Ommer.
\newblock Taming transformers for high-resolution image synthesis.
\newblock In \emph{Proceedings of the IEEE/CVF conference on computer vision and pattern recognition}, pages 12873--12883, 2021.

\bibitem[Gray(1984)]{gray1984vector}
Robert Gray.
\newblock Vector quantization.
\newblock \emph{IEEE Assp Magazine}, 1\penalty0 (2):\penalty0 4--29, 1984.

\bibitem[Gu and Dao(2023)]{gu2023mamba}
Albert Gu and Tri Dao.
\newblock Mamba: Linear-time sequence modeling with selective state spaces.
\newblock \emph{arXiv preprint arXiv:2312.00752}, 2023.

\bibitem[Gupta et~al.(2023)Gupta, Yu, Sohn, Gu, Hahn, Fei-Fei, Essa, Jiang, and Lezama]{gupta2023photorealistic}
Agrim Gupta, Lijun Yu, Kihyuk Sohn, Xiuye Gu, Meera Hahn, Li Fei-Fei, Irfan Essa, Lu Jiang, and Jos{\'e} Lezama.
\newblock Photorealistic video generation with diffusion models.
\newblock \emph{arXiv preprint arXiv:2312.06662}, 2023.

\bibitem[Huang et~al.(2023)Huang, Mao, Chen, and Zhang]{huang2023towards}
Mengqi Huang, Zhendong Mao, Zhuowei Chen, and Yongdong Zhang.
\newblock Towards accurate image coding: Improved autoregressive image generation with dynamic vector quantization.
\newblock In \emph{Proceedings of the IEEE/CVF Conference on Computer Vision and Pattern Recognition}, pages 22596--22605, 2023.

\bibitem[Isola et~al.(2017)Isola, Zhu, Zhou, and Efros]{isola2017image}
Phillip Isola, Jun-Yan Zhu, Tinghui Zhou, and Alexei~A Efros.
\newblock Image-to-image translation with conditional adversarial networks.
\newblock In \emph{Proceedings of the IEEE conference on computer vision and pattern recognition}, pages 1125--1134, 2017.

\bibitem[Kingma and Ba(2014)]{kingma2014adam}
Diederik~P Kingma and Jimmy Ba.
\newblock Adam: A method for stochastic optimization.
\newblock \emph{arXiv preprint arXiv:1412.6980}, 2014.

\bibitem[Lee et~al.(2022)Lee, Kim, Kim, Cho, and Han]{lee2022autoregressive}
Doyup Lee, Chiheon Kim, Saehoon Kim, Minsu Cho, and Wook-Shin Han.
\newblock Autoregressive image generation using residual quantization.
\newblock In \emph{Proceedings of the IEEE/CVF Conference on Computer Vision and Pattern Recognition}, pages 11523--11532, 2022.

\bibitem[Mentzer et~al.(2023)Mentzer, Minnen, Agustsson, and Tschannen]{mentzer2023finite}
Fabian Mentzer, David Minnen, Eirikur Agustsson, and Michael Tschannen.
\newblock Finite scalar quantization: Vq-vae made simple.
\newblock \emph{arXiv preprint arXiv:2309.15505}, 2023.

\bibitem[Niklaus and Liu(2020)]{Niklaus_CVPR_2020}
Simon Niklaus and Feng Liu.
\newblock Softmax splatting for video frame interpolation.
\newblock In \emph{IEEE Conference on Computer Vision and Pattern Recognition}, 2020.

\bibitem[Pont-Tuset et~al.(2017)Pont-Tuset, Perazzi, Caelles, Arbel{\'a}ez, Sorkine-Hornung, and Van~Gool]{pont20172017}
Jordi Pont-Tuset, Federico Perazzi, Sergi Caelles, Pablo Arbel{\'a}ez, Alex Sorkine-Hornung, and Luc Van~Gool.
\newblock The 2017 davis challenge on video object segmentation.
\newblock \emph{arXiv preprint arXiv:1704.00675}, 2017.

\bibitem[Press et~al.(2021)Press, Smith, and Lewis]{press2021train}
Ofir Press, Noah~A Smith, and Mike Lewis.
\newblock Train short, test long: Attention with linear biases enables input length extrapolation.
\newblock \emph{arXiv preprint arXiv:2108.12409}, 2021.

\bibitem[Skorokhodov et~al.(2022)Skorokhodov, Tulyakov, and Elhoseiny]{skorokhodov2022stylegan}
Ivan Skorokhodov, Sergey Tulyakov, and Mohamed Elhoseiny.
\newblock Stylegan-v: A continuous video generator with the price, image quality and perks of stylegan2.
\newblock In \emph{Proceedings of the IEEE/CVF conference on computer vision and pattern recognition}, pages 3626--3636, 2022.

\bibitem[Soomro(2012)]{soomro2012ucf101}
K Soomro.
\newblock Ucf101: A dataset of 101 human actions classes from videos in the wild.
\newblock \emph{arXiv preprint arXiv:1212.0402}, 2012.

\bibitem[Su et~al.(2024)Su, Ahmed, Lu, Pan, Bo, and Liu]{su2024roformer}
Jianlin Su, Murtadha Ahmed, Yu Lu, Shengfeng Pan, Wen Bo, and Yunfeng Liu.
\newblock Roformer: Enhanced transformer with rotary position embedding.
\newblock \emph{Neurocomputing}, 568:\penalty0 127063, 2024.

\bibitem[Tian et~al.(2024)Tian, Jiang, Yuan, Peng, and Wang]{tian2024visual}
Keyu Tian, Yi Jiang, Zehuan Yuan, Bingyue Peng, and Liwei Wang.
\newblock Visual autoregressive modeling: Scalable image generation via next-scale prediction.
\newblock \emph{arXiv preprint arXiv:2404.02905}, 2024.

\bibitem[Unterthiner et~al.(2018)Unterthiner, Van~Steenkiste, Kurach, Marinier, Michalski, and Gelly]{unterthiner2018towards}
Thomas Unterthiner, Sjoerd Van~Steenkiste, Karol Kurach, Raphael Marinier, Marcin Michalski, and Sylvain Gelly.
\newblock Towards accurate generative models of video: A new metric \& challenges.
\newblock \emph{arXiv preprint arXiv:1812.01717}, 2018.

\bibitem[Van Den~Oord et~al.(2017)Van Den~Oord, Vinyals, et~al.]{van2017neural}
Aaron Van Den~Oord, Oriol Vinyals, et~al.
\newblock Neural discrete representation learning.
\newblock \emph{Advances in neural information processing systems}, 30, 2017.

\bibitem[Vaswani(2017)]{vaswani2017attention}
A Vaswani.
\newblock Attention is all you need.
\newblock \emph{Advances in Neural Information Processing Systems}, 2017.

\bibitem[Villegas et~al.(2022)Villegas, Babaeizadeh, Kindermans, Moraldo, Zhang, Saffar, Castro, Kunze, and Erhan]{villegas2022phenaki}
Ruben Villegas, Mohammad Babaeizadeh, Pieter-Jan Kindermans, Hernan Moraldo, Han Zhang, Mohammad~Taghi Saffar, Santiago Castro, Julius Kunze, and Dumitru Erhan.
\newblock Phenaki: Variable length video generation from open domain textual descriptions.
\newblock In \emph{International Conference on Learning Representations}, 2022.

\bibitem[Wang et~al.(2024)Wang, Jiang, Yuan, Peng, Wu, and Jiang]{wang2024omnitokenizer}
Junke Wang, Yi Jiang, Zehuan Yuan, Binyue Peng, Zuxuan Wu, and Yu-Gang Jiang.
\newblock Omnitokenizer: A joint image-video tokenizer for visual generation.
\newblock \emph{arXiv preprint arXiv:2406.09399}, 2024.

\bibitem[Yan et~al.(2021)Yan, Zhang, Abbeel, and Srinivas]{yan2021videogpt}
Wilson Yan, Yunzhi Zhang, Pieter Abbeel, and Aravind Srinivas.
\newblock Videogpt: Video generation using vq-vae and transformers.
\newblock \emph{arXiv preprint arXiv:2104.10157}, 2021.

\bibitem[Yu et~al.(2021)Yu, Li, Koh, Zhang, Pang, Qin, Ku, Xu, Baldridge, and Wu]{yu2021vector}
Jiahui Yu, Xin Li, Jing~Yu Koh, Han Zhang, Ruoming Pang, James Qin, Alexander Ku, Yuanzhong Xu, Jason Baldridge, and Yonghui Wu.
\newblock Vector-quantized image modeling with improved vqgan.
\newblock \emph{arXiv preprint arXiv:2110.04627}, 2021.

\bibitem[Yu et~al.(2023{\natexlab{a}})Yu, Lezama, Gundavarapu, Versari, Sohn, Minnen, Cheng, Gupta, Gu, Hauptmann, et~al.]{yu2023language}
Lijun Yu, Jos{\'e} Lezama, Nitesh~B Gundavarapu, Luca Versari, Kihyuk Sohn, David Minnen, Yong Cheng, Agrim Gupta, Xiuye Gu, Alexander~G Hauptmann, et~al.
\newblock Language model beats diffusion--tokenizer is key to visual generation.
\newblock \emph{arXiv preprint arXiv:2310.05737}, 2023{\natexlab{a}}.

\bibitem[Yu et~al.(2023{\natexlab{b}})Yu, Sohn, Kim, and Shin]{PVDM}
Sihyun Yu, Kihyuk Sohn, Subin Kim, and Jinwoo Shin.
\newblock Video probabilistic diffusion models in projected latent space, 2023{\natexlab{b}}.

\bibitem[Zhang et~al.(2020)Zhang, Xu, Liu, Wang, Wu, Liu, and Jiang]{dtvnet}
Jiangning Zhang, Chao Xu, Liang Liu, Mengmeng Wang, Xia Wu, Yong Liu, and Yunliang Jiang.
\newblock Dtvnet: Dynamic time-lapse video generation via single still image.
\newblock In \emph{European Conference on Computer Vision}, pages 300--315. Springer, 2020.

\bibitem[Zhang et~al.(2018)Zhang, Isola, Efros, Shechtman, and Wang]{zhang2018unreasonable}
Richard Zhang, Phillip Isola, Alexei~A Efros, Eli Shechtman, and Oliver Wang.
\newblock The unreasonable effectiveness of deep features as a perceptual metric.
\newblock In \emph{Proceedings of the IEEE conference on computer vision and pattern recognition}, pages 586--595, 2018.

\bibitem[Zhao et~al.(2024)Zhao, Xiong, and Kr{\"a}henb{\"u}hl]{zhao2024image}
Yue Zhao, Yuanjun Xiong, and Philipp Kr{\"a}henb{\"u}hl.
\newblock Image and video tokenization with binary spherical quantization.
\newblock \emph{arXiv preprint arXiv:2406.07548}, 2024.

\end{thebibliography}

}
\end{document}